# Deep Learning Transformer Architecture for Named-Entity Recognition on Low-Resourced Languages: State of the art results

Ridewaan Hanslo
University of Pretoria
Gauteng, South Africa
Email: ridewaan.hanslo@up.ac.za

*Abstract*—This paper reports on the evaluation of Deep Learning (DL) transformer architecture models for Named-Entity Recognition (NER) on ten low-resourced South African (SA) languages. In addition, these DL transformer models were compared to other Neural Network and Machine Learning (ML) NER models. The findings show that transformer models substantially improve performance when applying discrete fine-tuning parameters per language. Furthermore, fine-tuned transformer models outperform other neural network and machine learning models on NER with the low-resourced SA languages. For example, the transformer models obtained the highest F-scores for six of the ten SA languages and the highest average F-score surpassing the Conditional Random Fields ML model. Practical implications include developing high-performance NER capability with less effort and resource costs, potentially improving downstream NLP tasks such as Machine Translation (MT). Therefore, the application of DL transformer architecture models for NLP NER sequence tagging tasks on low-resourced SA languages is viable. Additional research could evaluate the more recent transformer architecture models on other Natural Language Processing tasks and applications, such as Phrase chunking, MT, and Part-of-Speech tagging.

## I. Introduction

NATURAL Language Processing (NLP) which has been in existence for more than 70 years, is a branch of Artificial Intelligence [7]. NLP uses computational techniques for the analysis and representation of naturally occurring texts to achieve human-like language processing for various applications and tasks. Machine Translation (MT) was the first computer-based NLP application [7]. Thereafter, applications utilizing NLP such as Information Retrieval, Information Extraction (IE), and Question-Answering (QA) were introduced [7]. These IE applications include sequence tagging tasks such as Named-Entity Recognition (NER) and Part-of-Speech (POS) tagging.

NER is a task that processes natural language, classifying and grouping, for example, words into categories (also known as phrase types) [20]. With the advent of big data and large datasets, classifying natural language in these datasets has become increasingly important. For example, organizations are able to apply NER in customer support, content classification, and search and recommendation engines [21]. Furthermore, NER findings may be transferred to other NLP tasks such as MT, automatic text summarization, and knowledge base construction [20]. Lack of data severely impedes performance on NER tasks with low-resourced languages [20].

Recently, within NLP research, the use of Neural Network (NN) architectures, also referred to as Deep Learning (DL) architectures, has generated state-of-the-art results for MT, IE, and QA tasks [2], [8]. NN has seen several additions in the past few decades, from Convolutional Neural Networks (CNN) and Recurrent Neural Networks (RNN) to Transformer architectures [3], [4], [8]. CNN is an extensively studied DL architecture inspired by the visual perception mechanisms of living creatures [11]. RNN is concerned with sequential data that display correlations between data points within a time sequence [12]. Transformers are prominent NN architectures in NLP research, surpassing RNN and CNN in model performance [13].

Transformers facilitate the creation of high-capacity models that are pre-trained on large corpora. These transformers capture long-range sequence features that facilitate parallel training, and the pre-trained models are easily adapted to specific tasks with good performance [13]. XLM-Roberta (XLM-R) is a recent transformer model that has reported state-of-the-art results for NLP tasks and applications, such as NER, POS tagging, phrase chunking, and MT [2], [9].

NLP sequence tagging tasks such as NER and POS tagging have been extensively researched [1]-[7], [9], [10]. However, within the past few years, new DL transformer architecture models such as XLM-R, Multilingual Bidirectional Encoder Representations from Transformers (M-BERT), and Cross-Lingual Language Model (XLM) lower the time needed to train large datasets through greater parallelization. This allows low-resourced languages to be trained and tested with less effort and resource costs while achieving state-of-the-art results for sequence tagging tasks [1], [2], [9]. M-BERT, a single language model pre-trained from monolingual corpora, performs cross-lingual generalization very well [14]. Furthermore, M-BERT is capable of capturing multilingual representations [14]. On the other hand, XLM pre-training has led to solid

This work is based on the research supported by the National Research Foundation of South Africa (Grant Number 138325).

improvements in NLP benchmarks [15]. Additionally, XLM models have contributed to significant improvements in NLP studies involving low-resourced languages [15]. These transformer models are usually trained on very large corpora with datasets that are terabytes (TB) in size.

## II. BACKGROUND

A recent study by [1] researched whether NN's are viable for NLP sequence tagging (POS tagging and NER) and sequence-to-sequence (Lemmatization and Compound Analysis) tasks for resource-scarce languages. These resource-scarce languages are ten of the 11 official South African (SA) languages, with English being excluded. The languages are considered low-resourced, with Afrikaans (af) being the more resourced of the ten [1], [10]. This recent study compared two Bidirectional Long Short-Term Memory with Auxiliary Loss (bi-LSTM-aux) NN models to a baseline Conditional Random Fields (CRF) model. The annotated data used for the experiments are derived from the National Centre for Human Language Technology (NCHLT) text project. The results suggest that NN architectures such as bi-LSTM-aux are viable for sequence tagging tasks for most SA languages [1]. However, within the study by [1], NN's did not outperform the CRF Machine Learning (ML) baseline NER model. Rather the CRF model performed better on NER than the bi-LSTM-aux models. Loubser and Puttkammer [1], therefore, advised further studies to be conducted using NN transformer models on resource-scarce SA languages. Additionally, because of the considerable variation in performance per language during their study, [1] suggests conducting further research on the variation in performance per language.

Similarly, a previous study by [18] evaluated XLM-R transformer models for NER on low-resourced languages. However, the fine-tuning of the transformer models was at the model level and not the language level. In other words, a transformer model was fined-tuned on, for example, the Afrikaans (af) language. Thereafter, the model with the fine-tuned parameters for the Afrikaans (af) language was applied to the other remaining nine SA languages. The reason for this decision was due to resource capacity constraints. As a result, the study by [18] produced only a couple higher F-scores than the CRF and bi-LSTM-aux baseline models. Albeit, the CRF model retained the highest average F-score for the ten languages.

For this reason, this study builds upon these previous studies by evaluating the performance of DL transformer architecture for NER on low-resourced languages with fine-tuning of the model applied at the language level. Therefore, the purpose of this study is to evaluate the performance of the NLP NER sequential task using two XLM-R transformer models applying fine-tuning to each model and language combination. In addition, the experiment results are compared to previous research findings.

### A. Research Hypotheses

**H₁** – There is a performance improvement for NER on the low-resourced SA languages using fine-tuned XLM-R transformer models.

**H₂** – Fine-tuned XLM-R transformer models outperform other neural network and machine learning models on NER with the low-resourced SA languages.

### B. Paper Layout

The remainder of this paper comprises of the following sections: Sect. III provides information on the languages and datasets while Sect. IV presents the language model architectures. The experiment settings are presented in Sect. V. The results and a discussion of the research findings are provided in Sect. VI and Sect. VII respectively. Section VIII concludes the paper, providing practical implications, limitations of this study and recommendations for future research.

## III. LANGUAGES AND DATASETS

As mentioned by [1], SA is a country with at least 35 spoken languages. Of those languages, 11 are granted official status. The 11 languages can further be broken up into three distinct groups. The two West-Germanic languages, English and Afrikaans (af). Five disjunctive languages, Tshivenda (ve), Xitsonga (ts), Sesotho (st), Sepedi (nso) and Setswana (tn) and four conjunctive languages, isiZulu (zu), isiXhosa (xh), isiNdebele (nr) and Siswati (ss). A difference between SA disjunctive and conjunctive languages is the former has more words per sentence than the latter. Therefore, disjunctive languages have a higher token count than conjunctive languages. For further details on conjunctive and disjunctive languages and examples thereof, see [1].

The datasets for the ten evaluated languages are available from the South African Centre for Digital Language Resources online repository (https://repo.sadilar.org/). These annotated datasets are part of the NCHLT Text Resource Development Project, developed by the Centre for Text Technology (CTexT, North-West University, South Africa) with contributions by the SA Department of Arts and Culture. The annotated data is tokenized into five phrase types. These five phrase types are:

1. LOC - Location
2. MISC - Miscellaneous
3. ORG - Organization
4. OUT - not considered part of any named-entity
5. PER - Person

The LOC, ORG and PER phrase types are entity names and are the main named entity category used in this study. The MISC phrase type as explained by [10] are for phrase types that form part of either the number expressions or temporal named entity categories so as not to lose the opportunity to annotate the data, which can be further annotated in the future [10].

It is important to note that this annotated data is the same dataset used by the previous studies [1], [10], [18]. However, the studies by [10] and [18] clearly indicates the inclusion of the MISC phrase type whereas, the study by [1] does not.

The previous studies made use of the CoNLL-2003 shared task protocol for data tagging [19]. Additionally, the named entities are further annotated with the beginning [B], inside [I], and outside [O] labelling scheme, which is posited to be ideal for sequence tagging training [10].

The datasets consist of SA government domain corpora. Therefore, the SA government domain corpora are used to do the experiments and comparisons. Eiselen [10] provides further details on the annotated corpora.

## IV. LANGUAGE MODEL ARCHITECTURES

### A. XLM-R

XLM-Roberta (XLM-R) is a transformer-based multilingual masked language model [2]. This language model trained on 100 languages uses 2.5 TB of CommonCrawl (CC) data [2]. From the 100 languages used by the XLM-R multilingual masked language model, it is noted that Afrikaans (af) and isiXhosa (xh) are included in the pre-training.

As indicated by [2], the benefit of this model is training the XLM-R model on cleaned CC data increases the amount of data for low-resourced languages. Further, because the XLM-R multilingual model is pre-trained on many languages, low-resourced languages improve performance due to positive transfer [2].

The study by [2] reports the state-of-the-art XLM-R model performs better than other NN models such as mBERT and XLM on QA, classification, and sequence labelling. For this research study, two transformer models are used for NER evaluation. The $XLM-R_{Base}$ NN model and the $XLM-R_{Large}$ NN model. The $XLM-R_{Base}$ model has 12 layers, 768 hidden states, 12 attention heads, 250 thousand vocabulary size, and 270 million parameters. The $XLM-R_{Large}$ model has 24 layers, 1024 hidden states, 16 attention heads, 250 thousand vocabulary size, and 550 million parameters [2]. Both pre-trained models are publicly available (https://bit.ly/xlm-rbase, https://bit.ly/xlm-rlarge).

These pre-trained models ($XLM-R_{Base}$ and $XLM-R_{Large}$), as mentioned in Sect. I allow low-resourced languages to be trained and tested with fewer resource costs and effort. Therefore, they were fed into this study's DL transformer architecture NER model as part of the NER evaluation process. The model was developed with the Python programming language, the PyTorch ML framework, the Facebook AI Research Sequence-to-Sequence Toolkit (written in Python), and the PyTorch Transformers library. The developed model incorporated the AdamW PyTorch algorithm (optimizer) with warm-up scheduling was trained, validated and then evaluated on the test data. Section V discusses the model's experimental settings.

### B. CRF

Conditional Random Fields (CRF) is used for building probabilistic models for segmentation and labelling of sequence data [5]. CRF as ML models are simple, yet, successfully used for NLP sequence tagging tasks, such as NER and POS tagging [4]. Before the use of CRF, Hidden Markov Models (HMM) and stochastic grammars were widely used probabilistic models for tagging tasks [5]. The benefit of using CRF as a sequence modelling framework is it addresses label biases much better than HMM [5]. Additionally, CRF also provides for better stochastic context-free grammar generalization. More information on the CRF ML model is provided by [5].

Eiselen [10] used a CRF ML model for NER on the ten low-resourced SA languages, and [1] included this model as the baseline to compare their two bi-LSTM-aux NN architecture models. Loubser and Puttkammer's [1] findings show that the bi-LSTM-aux models were almost on par with the CRF model, meaning that the NN models did not outperform the ML model. For this reason, this study includes the CRF model as it would be good to compare the DL models with ML models that consistently performs well on NLP sequence tasks. The code for this model is publicly available (https://taku910.github.io/crfpp).

### C. bi-LSTM-aux

Bidirectional Long Short-Term Memory with Auxiliary Loss (bi-LSTM-aux) NN models have been reported as successful with NLP sequence modelling tasks [3]. Modelling tasks include POS tagging, NER, sentiment analysis, and dependency parsing [3]. While both LSTM and bi-LSTM models are classified as RNN, bi-LSTMs implement a backward and forward pass through the sequence before proceeding to the next layer within the network [3]. The inclusion of the auxiliary loss function in the model is to help improve performance gains for rare words used within the corpora [3]. The bi-LSTM-aux model was trained on 22 languages, using polyglot embeddings, and data obtained from the Universal Dependencies project [3]. Additional details on the model are obtainable from [3]. The findings from the study by [3] using their novel bi-LSTM-aux model for POS tagging suggests that the model is as effective as HMM and CRF tagging models. Loubser and Puttkammer [1] used bi-LSTM-aux and bi-LSTM-aux with embeddings model variations for their study. The code for these models is publicly available (https://github.com/bplank/bilstm-aux).

## V. EXPERIMENTAL SETTINGS

The experimental settings for the $XLM-R_{Base}$ and $XLM-R_{Large}$ models are described next, followed by the evaluation metrics and the corpora descriptive statistics.

TABLE I.
FINE-TUNED PARAMETERS PER LANGUAGE AND MODEL COMBINATION

| Language | Model | Learning Rate | Warmup Proportion | Dropout |
|---|---|---|---|---|
| Afrikaans (af) | Base | 6e-5 | 0.0 | 0.0 |
| | Large | 6e-5 | 0.0 | 0.2 |
| isiNdebele (nr) | Base | 6e-5 | 0.0 | 0.0 |
| | Large | 6e-5 | 0.0 | 0.0 |
| isiXhosa (xh) | Base | 6e-5 | 0.0 | 0.2 |
| | Large | 6e-5 | 0.1 | 0.2 |
| isiZulu (zu) | Base | 6e-5 | 0.1 | 0.2 |
| | Large | 6e-5 | 0.0 | 0.2 |
| Sepedi (nso) | Base | 7e-5 | 0.1 | 0.3 |
| | Large | 7e-5 | 0.1 | 0.3 |
| Sesotho (st) | Base | 6e-5 | 0.0 | 0.2 |
| | Large | 6e-5 | 0.1 | 0.2 |
| Setswana (tn) | Base | 6e-5 | 0.1 | 0.3 |
| | Large | 6e-5 | 0.0 | 0.0 |
| Siswati (ss) | Base | 6e-5 | 0.0 | 0.2 |
| | Large | 6e-5 | 0.0 | 0.2 |
| Tshivenda (ve) | Base | 6e-5 | 0.0 | 0.2 |
| | Large | 6e-5 | 0.0 | 0.2 |
| Xitsonga (ts) | Base | 6e-5 | 0.0 | 0.2 |
| | Large | 6e-5 | 0.0 | 0.0 |

### A. XLM-R Settings

The training, validation, and test dataset split was 80%, 10%, and 10%, respectively. Table I provides the fine-tuned parameters at the model and language level while the shared settings across the models and languages are as follows:
- Gradient accumulation steps: 4
- Maximum sequence length: 128
- Training batch size: 32
- Training epochs: 10

### B. Evaluation Metrics

Precision, Recall and F-score are evaluation metrics used for text classification tasks, such as NER. These metrics are used to measure the model's performance during the experiments. The formulas for these metrics leave out the correct classification of true negatives (*tn*) and false negatives (*fn*), referred to as negative examples, with greater importance placed on the correct classification of positive examples such as true positives (*tp*) and false positives (*fp*) [16]. For example, correctly classified spam emails (*tp*) are more important than correctly classified non-spam emails (*tn*). In addition, multi-class classification was used for the research experiments to classify a token into a discrete class from three or more classes. The metric's macro-averages were used for evaluation and comparison. Macro-averaging (*M*) treats classes equally, while micro-averaging (*μ*) favors bigger classes [16]. Each evaluation metric and its formula as described by [16] are listed below.

Precision$_M$: "*the number of correctly classified positive examples divided by the number of examples labeled by the system as positive*" (1).

$$\frac{\sum_{i=1}^{l} \frac{tp_i}{tp_i + fp_i}}{l} \quad (1)$$

Recall$_M$: "*the number of correctly classified positive examples divided by the number of positive examples in the data*" (2).

$$\frac{\sum_{i=1}^{l} \frac{tp_i}{tp_i + fn_i}}{l} \quad (2)$$

Fscore$_M$: "*a combination of the above*" (3).

$$\frac{(\beta^2 + 1) Precision_M Recall_M}{\beta^2 \, Precision_M + Recall_M} \quad (3)$$

### C. Corpora Descriptive Statistics

Table II provides descriptive statistics for the language's training data. As mentioned earlier, disjunctive languages have a higher token count than conjunctive languages. Albeit, the unique phrase type and named entity count for conjunctive languages are, on average, higher than the disjunctive languages.

TABLE II.
THE TEN LANGUAGES TRAINING DATA DESCRIPTIVE STATISTICS

| Language | Writing System | Tokens | Phrase Types | Named Entities |
|---|---|---|---|---|
| af | Mixed | 184 005 | 22 693 | 21 100 |
| nr | Conjunctive | 129 577 | 38 852 | 25 030 |
| xh | Conjunctive | 96 877 | 33 951 | 15 185 |
| zu | Conjunctive | 161 497 | 50 114 | 25 216 |
| nso | Disjunctive | 161 161 | 17 646 | 19 163 |
| st | Disjunctive | 215 655 | 18 411 | 19 211 |
| tn | Disjunctive | 185 433 | 17 670 | 18 993 |
| ss | Conjunctive | 140 783 | 42 111 | 21 403 |
| ve | Disjunctive | 188 399 | 15 947 | 14 119 |
| ts | Disjunctive | 214 835 | 17 904 | 24 376 |

## VI. RESULTS

Fig. 1 depicts the precision scores for the ten low-resourced SA languages under the five NER models. The Afrikaans (af) language has the highest precision score with 81.74% for the XLM-R$_{Large}$ model, while the Sesotho (st) language has the lowest precision score of 50.31% for the bi-LSTM-aux emb model. Table III displays the precision scores of the two XLM-R transformer models compared to models used by [1] and [10]. The XLM-R models share five of the ten highest precision scores, with the highest average score belonging to the CRF model with 75.64%. The bold scores in Tables III, IV, V and VI show the highest evaluation metric score for each language and the model with the highest average score.

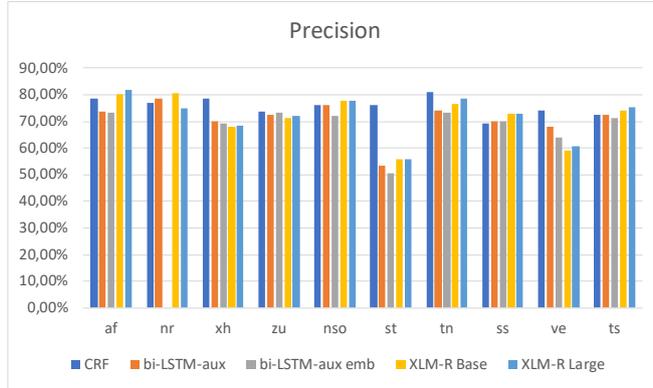

Fig. 1 The precision % for the 10 low-resourced SA languages visual representation

Fig. 2 depicts the recall scores for the ten SA languages under the five NER models. As with the precision evaluation metric, the Afrikaans (af) language has the highest recall scores for three of the five models, with an 87.07% for the XLM-R$_{Large}$ model. Sesotho (st) has the lowest recall score of 55.56% for the bi-LSTM-aux model. Table IV displays the recall scores for the ten low-resourced SA languages. The XLM-R models share the highest recall scores for seven of the ten languages, with the highest average score belonging to the XLM-R$_{Base}$ model with 76.34%.

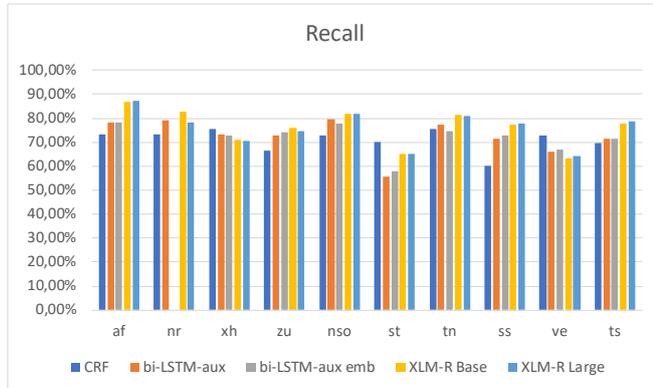

Fig. 2 The recall % for the 10 low-resourced SA languages visual representation

Fig. 3 depicts the F-scores for the ten languages under the five NER models. The Afrikaans (af) language has the highest F-scores for three of the five models, with 84.25% for the XLM-R$_{Large}$ model. Sesotho (st) has the lowest F-score of 53.77% for the bi-LSTM-aux emb model. Table V displays the F-score comparison. The XLM-R models produced the highest F-scores for six of the ten languages, with the highest average score belonging to the XLM-R$_{Base}$ model with 73.64%.

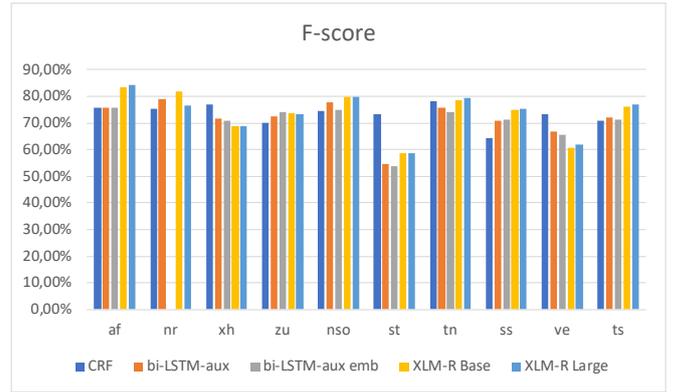

Fig. 3 The F-score % for the 10 low-resourced SA languages visual representation

## VII. DISCUSSION

This section discusses the research findings concerning hypotheses testing. The first alternate hypothesis is accepted or rejected based on the XLM-R transformer model's performance using the three-evaluation metrics. The second hypothesis is accepted or rejected based on the XLM-R transformer model's performance compared to the CRF and bi-LSTM-aux models used in previous SA NER studies.

$H_1$ – There is a performance improvement for NER on the low-resourced SA languages using fine-tuned XLM-R transformer models.

The XLM-R$_{Large}$ and XLM-R$_{Base}$ transformer models produced F-scores that ranged from 53.77% for the Sesotho (st) language to 84.25% for the Afrikaans (af) language. In addition, many of the models recall scores were greater than 75% whereas, the precision scores were averaging at 70%. Remember, in this instance, the recall metric emphasizes the average per-named-entity effectiveness of the classifier to identify named entities, and the precision metric compares the alignment of the classifier's average per-named-entities to the named entities in the data. All F-scores were above 60% except the Sesotho (st) language, which for both XLM-R models were below 60%.

A previous study by [18] was only able to achieve F-scores of 39% for the Sesotho (st) language and proposed that using different hyper-parameter tuning (fine-tuning) and dataset splits could produce higher F-scores. Further, [18] also suggested that further studies could implement the transformer models with discrete fine-tuning parameters per language to produce higher F-scores. The findings of this study show that transformer models with discrete fine-tuning parameters per language generate higher F-scores (see Table VI). The fine-tuned transformer models produced an average F-score 6% higher than the previous transformer models.

TABLE III.
THE PRECISION % COMPARISON BETWEEN TRANSFORMER MODELS AND PREVIOUS SA LANGUAGE NER STUDIES

| | Precision | | | | |
|---|---|---|---|---|---|
| | CRF* | bi-LSTM-aux** | bi-LSTM-aux emb** | XLM-R$_{Base}$ | XLM-R$_{Large}$ |
| af | 78.59% | 73.61% | 73.41% | 80.35% | **81.74%** |
| nr | 77.03% | 78.58% | n/a*** | **80.74%** | 74.73% |
| xh | **78.60%** | 69.83% | 69.08% | 67.94% | 68.46% |
| zu | **73.56%** | 72.43% | 73.44% | 71.26% | 71.91% |
| nso | 76.12% | 75.91% | 72.14% | **77.90%** | 77.75% |
| st | **76.17%** | 53.29% | 50.31% | 55.67% | 55.62% |
| tn | **80.86%** | 74.14% | 73.45% | 76.58% | 78.65% |
| ss | 69.03% | 70.02% | 69.93% | **72.98%** | 72.84% |
| ve | **73.96%** | 67.97% | 63.82% | 58.85% | 60.61% |
| ts | 72.48% | 72.33% | 71.03% | 74.18% | **75.15%** |
| Average | **75.64%** | 70.81% | 68.51% | 71.64% | 71.74% |

\* As reported by [10]. ** As reported by [1]. *** No embeddings were available for isiNdebele.

TABLE IV.
THE RECALL % COMPARISON BETWEEN TRANSFORMER MODELS AND PREVIOUS SA LANGUAGE NER STUDIES

| | Recall | | | | |
|---|---|---|---|---|---|
| | CRF* | bi-LSTM-aux** | bi-LSTM-aux emb** | XLM-R$_{Base}$ | XLM-R$_{Large}$ |
| af | 73.32% | 78.23% | 78.23% | 86.89% | **87.07%** |
| nr | 73.26% | 79.20% | n/a*** | **82.92%** | 78.27% |
| xh | **75.61%** | 73.30% | 72.78% | 71.05% | 70.48% |
| zu | 66.64% | 72.64% | 74.32% | **75.92%** | 74.58% |
| nso | 72.88% | 79.66% | 77.63% | 81.85% | **82.05%** |
| st | **70.27%** | 55.56% | 57.73% | 65.04% | 65.04% |
| tn | 75.47% | 77.42% | 74.71% | **81.38%** | 80.74% |
| ss | 60.17% | 71.44% | 72.82% | 77.20% | **77.88%** |
| ve | **72.92%** | 65.91% | 67.09% | 63.24% | 64.22% |
| ts | 69.46% | 71.44% | 71.25% | 77.99% | **78.90%** |
| Average | 71.00% | 72.48% | 71.84% | **76.34%** | 75.92% |

\* As reported by [10]. ** As reported by [1]. *** No embeddings were available for isiNdebele.

TABLE V.
THE F-SCORE % COMPARISON BETWEEN TRANSFORMER MODELS AND PREVIOUS SA LANGUAGE NER STUDIES

| | F-score | | | | |
|---|---|---|---|---|---|
| | CRF* | bi-LSTM-aux** | bi-LSTM-aux emb** | XLM-R$_{Base}$ | XLM-R$_{Large}$ |
| af | 75.86% | 75.85% | 75.74% | 83.47% | **84.25%** |
| nr | 75.10% | 78.89% | n/a*** | **81.69%** | 76.44% |
| xh | **77.08%** | 71.52% | 70.88% | 68.85% | 68.80% |
| zu | 69.93% | 72.54% | **73.87%** | 73.48% | 73.17% |
| nso | 74.46% | 77.74% | 74.79% | 79.82% | **79.83%** |
| st | **73.09%** | 54.40% | 53.77% | 58.78% | 58.72% |
| tn | 78.06% | 75.74% | 74.07% | 78.70% | **79.54%** |
| ss | 64.29% | 70.72% | 71.35% | 74.91% | **75.19%** |
| ve | **73.43%** | 66.92% | 65.41% | 60.68% | 61.99% |
| ts | 70.93% | 71.88% | 71.14% | 76.03% | **76.97%** |
| Average | 73.22% | 71.62% | 70.11% | **73.64%** | 73.49% |

\* As reported by [10]. ** As reported by [1]. *** No embeddings were available for isiNdebele.

Therefore, the alternative hypothesis is accepted as there is a performance improvement on NER with the low-resourced SA languages using fine-tuned XLM-R transformer models. It is important to note that the previous study by [18] was not included in the comparative analysis with previous SA NER studies because the experimental results were insignificant when compared to the average F-scores of the [1] and [10] studies (see Table V).

**H₂** – Fine-tuned XLM-R transformer models outperform other neural network and machine learning models on NER with the low-resourced SA languages.

The fine-tuned transformer models (see Table I) were also compared to the findings of previous studies. In particular, [10] used the CRF ML model to do NER sequence tagging on the ten resource-scarce SA languages. Furthermore, [1] implemented bi-LSTM-aux NN models both with and without embeddings on the same datasets. The comparative

TABLE VI.
THE F-SCORE % COMPARISON BETWEEN TRANSFORMER MODELS AND FINE-TUNED TRANSFORMER MODELS

| | F-score | | | |
|---|---|---|---|---|
| | XLM-R$_{Base}$* | XLM-R$_{Large}$* | XLM-R$_{Base}$ | XLM-R$_{Large}$ |
| af | 82.47% | **84.25%** | 83.47% | **84.25%** |
| nr | 76.17% | 75.60% | **81.69%** | 76.44% |
| xh | 63.58% | 64.68% | **68.85%** | 68.80% |
| zu | 72.54% | 73.17% | **73.48%** | 73.17% |
| nso | 78.86% | n/a** | 79.82% | **79.83%** |
| st | 38.94% | 39.48% | **58.78%** | 58.72% |
| tn | 69.78% | 71.91% | 78.70% | **79.54%** |
| ss | 67.57% | 68.34% | 74.91% | **75.19%** |
| ve | 60.68% | **61.99%** | 60.68% | **61.99%** |
| ts | 65.57% | 66.12% | 76.03% | **76.97%** |
| Average | 67.61% | 67.28% | **73.64%** | 73.49% |

* As reported by [18]. ** The model was unable to produce scores for Sepedi.

analysis reveals the performance improvement of implementing DL transformer architecture for NLP sequence tagging tasks such as NER. For example, when analyzing the F-scores, the XLM-R models have the highest F-scores for six of the ten languages, and the CRF model has three of the highest F-scores (see Table V). Meanwhile, the bi-LSTM-aux models had only one of the highest F-scores (see Table V).

This study's result is an improvement for NER research in the SA context because the previous studies by [1] and [18] could not outperform the CRF ML model implemented by [10] until now. Albeit, not all the SA languages are good candidates for DL architectures. For example, the isiXhosa (xh) and Tshivenda (ve) languages consistently underperform compared to the CRF ML model. Additionally, the comparative analysis identified the Sesotho (st) language as the lowest-performing language across the NN models, with an average F-score of 56%, making it an outlier and an unviable language for current DL architectures for NER.

Therefore, the alternative hypothesis is accepted as fine-tuned XLM-R transformer models outperform other neural network and machine learning models on NER with the low-resourced SA languages (see Table V and Table VI).

This study reveals that the fine-tuned XLM-R transformer models perform relatively well on low-resourced SA languages with NER sequence tagging. Noticeably, there is no distinct performance difference between disjunctive and conjunctive languages. In addition, Afrikaans (af) outperform the other languages using the transformer models. As mentioned earlier, the outlier is the Sesotho (st) language, with the CRF baseline model F-score being 14% more than the XLM-R models and 18% more than the bi-LSTM-aux models. In confirmation with [18], including a language, such as isiXhosa (xh) during the transformer model pre-training does not guarantee good performance during evaluation.

Further, [10] suggested excluding the MISC phrase type to determine whether recall can be improved upon, however, this study revealed that even with the inclusion of the MISC type, the transformer models increased the recall scores considerably. Nonetheless, this is not to say that re-evaluation using an updated list of named entities will not produce higher metric scores.

VIII. CONCLUSION

This research reports on the implementation of Neural Network (NN) and Machine Learning (ML) models to evaluate Named-Entity Recognition (NER) sequence tagging on the ten low-resourced languages of South Africa (SA). The models were trained, validated, and tested using SA government domain corpora. Given the findings, the XLM-R transformer models performed better than the CRF and bi-LSTM models on recall and F-score. The transformer models produced higher F-scores for six of the ten SA languages, while the CRF model had only three of the highest F-scores. The CRF remained dominant on precision, averaging around 75%.

In addition, both the hypotheses were accepted (see Section VII). Firstly, using fine-tuned XLM-R transformer models improves performance on NER for low-resourced SA languages substantially. Secondly, fine-tuned XLM-R transformer models outperform other neural network and machine learning models on NER for the low-resourced SA languages. Therefore, NN transformer models are feasible for sequence tagging tasks, such as NER.

The implications of this study for research and practice of NER using NN transformer models are that such models are not only viable for low-resourced languages but advisable given they require less effort and resource costs. Furthermore, this approach to NER tasks benefits other downstream NLP tasks and applications. These tasks and applications include question answering, machine translation, and machine reading comprehension.

A limitation of this research is not evaluating the more recent XLM-R$_{XL}$ and XLM-R$_{XXL}$ models on the NER sequence tagging task. Furthermore, the datasets could be re-evaluated using an updated list of named entities.

Additional research could evaluate transformer models on other NLP applications and tasks. Further, NLP tasks and applications could be tested using a linear-complexity recurrent transformer variant and a frozen pre-trained transformer model.